\documentclass[prl,aps,twocolumn,superscriptaddress,showpacs]{revtex4}
\usepackage{amsmath}
\usepackage{amsbsy}
\usepackage{amssymb}
\usepackage[dvips]{graphicx}
\usepackage{texdraw}
\usepackage{pstricks}
\begin{document}

\title{Gaussian Belief with dynamic data and in dynamic network}

\author{Erik Aurell}
\affiliation{KTH Linnaeus Centre ACCESS, KTH-Royal Institute of Technology, Stockholm, Sweden}
\affiliation{Dept. Information and Computer Science, TKK-Helsinki University of Technology, Espoo, Finland}

\author{Ren\'{e} Pfitzner}
\affiliation{KTH Linnaeus Centre ACCESS, KTH-Royal Institute of Technology, Stockholm, Sweden}
\affiliation{Friedrich-Schiller-University Jena, Germany}

\date{\today}

\begin{abstract}
In this paper we analyse Belief Propagation over a Gaussian model in a dynamic environment.
Recently, this has been proposed as a method to average local measurement values by a distributed 
protocol (``Consensus Propagation'', Moallemi \& Van Roy, 2006), where the average is available 
for read-out at every single node. In the case that the underlying network is constant but the
values to be averaged fluctuate (``dynamic data''), convergence and accuracy are 
determined by the spectral properties of an associated Ruelle-Perron-Frobenius operator.
For Gaussian models on Erd\H{o}s-R\'{e}nyi graphs, numerical computation
points to a spectral gap remaining in the large-size limit, implying exceptionally 
good scalability. In a model where the underlying network 
also fluctuates (``dynamic network''), averaging is more effective than in
the dynamic data case. Altogether, this implies very good performance of these
methods in very large systems, and opens a new field of statistical physics
of large (and dynamic) information systems.  
\end{abstract}

\pacs{89.75.Hc, 89.90.+n}

\maketitle

\begin{figure}
\includegraphics[width=0.30\textwidth]{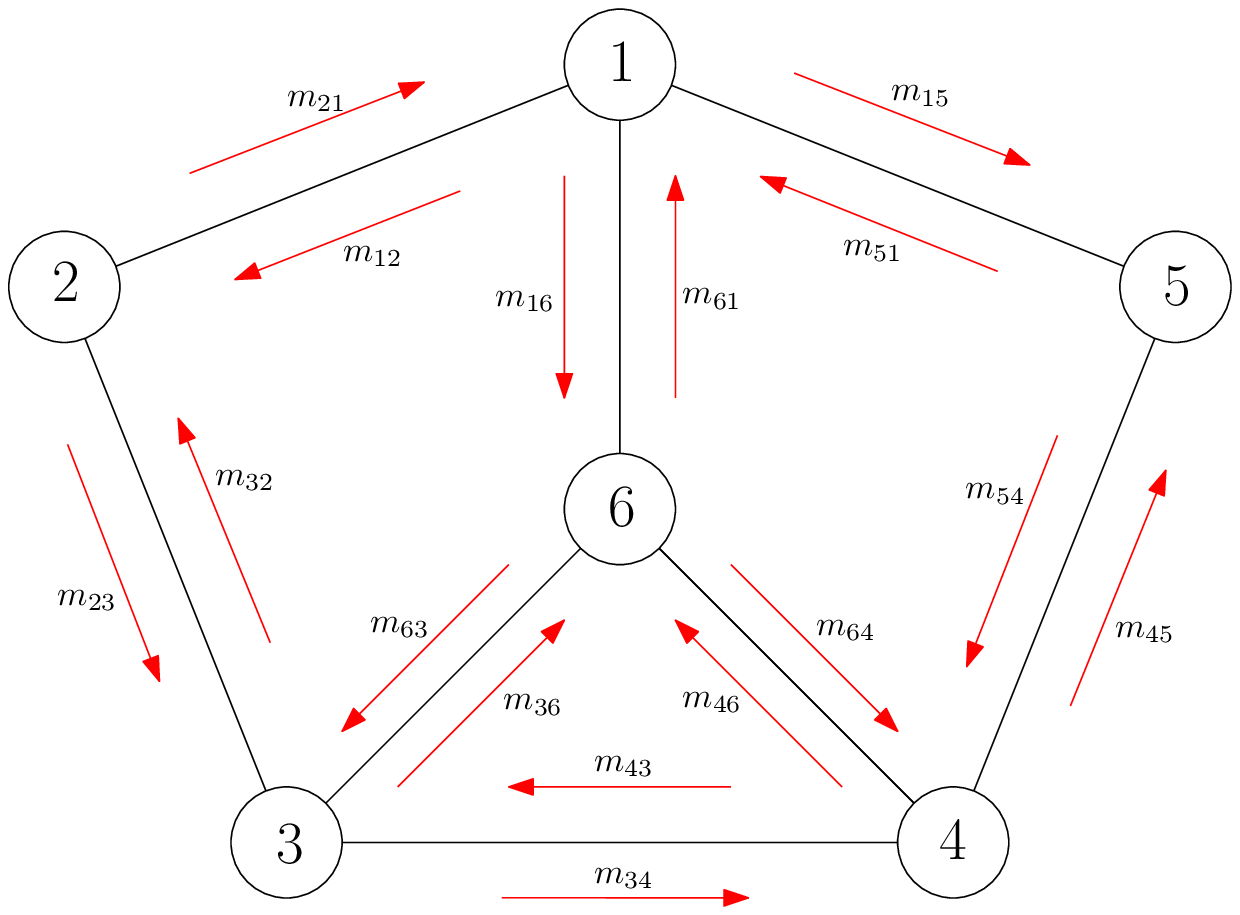}
\caption{Illustrated Belief Propagation message passing scheme in a 6-node network. In the Consensus Propagation case the messages $m_{ij}$ are decomposed into messages $K_{ij}$ and $\mu_{ij}$.}
\label{fig:network2}
\end{figure}

Message-passing algorithms have over the last two decades turned out to be an important paradigm in fields as distant as iterative decoding, image processing and AI, see~\cite{Pearl1988} for the intuition behind Belief Propagation (BP) in AI, and \cite{Cow1998,Kschischang2001,YedFreWei2003} for more recent reviews. It has been realized that systems where the message-passing algorithms are effective can often be assimilated to disordered systems in statistical physics, and that the message-passing algorithms themselves are closely related to the Bethe approximation~\cite{MezardMontanari2009}. Most applications pursued concern {\it inference} in {\it static models}; how to do this effectively (if approximately), and when these methods work. In another direction, Consensus Propagation (CP) has been proposed as a message-passing scheme to average measurement values in a network of connected nodes \cite{MoaVan2006}. This is a naturally {\it dynamic} setting, where, in large networks, and in many scenarios of interest, one must allow the measurement values, and perchance the network itself, to change on the same time scale as the averaging process. The two strands of inquiry are connected by the fact that CP is equivalent to BP on a class of Gauss-Markov random fields \cite{MoaVan2006,MJW2006}.
\\\\
In this contribution we study CP both in a static network with changing measurement values (dynamic data), and in a network where the strengths of the interconnections also change (dynamic network). We will show that the method has very good scalability, \textit{i.e.} that its performance degrades very slowly as the systems grow. In a sense, to be made precise below, performance does not degrade with size at all. This should make CP a very interesting method for aggregation tasks in large and dynamic networks, possibly competitive to alternative schemes such as gossiping~\cite{JeMoBa2005}. From a physics perspective the salient points are the following: (i) CP with dynamic data is (after a transient) a linear averaging process; (ii) the kernel of this averaging process, being the linearization of Gaussian BP, is related to the second variation of the Bethe free energy of the Gauss-Markov random field; (iii) the leading eigenvalue of the kernel is a self-averaging quantity in Erd\H{o}s-R\'{e}nyi networks, which in addition does not depend on the network size; (iv) CP with dynamic network and dynamic data functions as well (or better) as CB with dynamic data only. Points (ii) and (iii) imply that we identify a new random matrix construction with unexpected properties, and possibly important practical consequences. Point (iv) means concretely that dynamic data is the slow stable (and also flat) manifold of the kernel, while dynamic network spans the fast stable manifold. Perturbations in the dynamic network directions hence relax faster than dynamic data, which explains the good properties.
\\\\         
{\it Belief Propagation (BP) and Consensus Propagation (CP):}
BP is an algorithm to infer marginal probability distributions of a joint probability functions~\cite{YedFreWei2003}. It works via distributed message passing from each node of the underlying graph of the model to every neighboring node (FIG. \ref{fig:network2}). It is correct on tree-like graph topologies and has been shown to often yield good results in topologies including loops~\cite{MezardMontanari2009,MurWeiJo}. The messages in BP can be seen as 1-node marginal conditional probability distributions, which implies that BP works best computationally when the size of local state space is limited, e.g. for Ising spins. BP on Gauss-Markov random field is a special case, since Gaussianity is preserved under convolution, and the BP messages can be parametrized by two real values corresponding to (conditional) mean and (conditional) variance. A further very special property of BP on Gaussian models is that it is {\it exact} for the modes of the marginal distributions, in a very wide class of models~\cite{WeFree2001,MJW2006}.
\\\\
A special instance of Gaussian Belief Propagation is \textit{Consensus Propagation} \cite{MoaVan2006}. This algorithm aims to solve the problem of calculating the average $\overline{y}$ of some values $y_i$ (gathered by nodes $i$ in a network $G$) in a distributed way. The Gaussian model associated to CP is \cite{MoaVan2006}:
\begin{equation}
\label{eqn:CP}
p(x,\beta)=\frac{1}{Z}\exp(-\|x-y\|^2-\beta \sum_{{i,j} \in E}Q_{ij}(x_i-x_j)^2)
\end{equation}
In \eqref{eqn:CP}, $Z$ is a normalization, $\beta$ is a global and $Q_{ij}$ are local coupling parameters. 
BP on \eqref{eqn:CP} is guaranteed to converge for any finite $\beta$, and the modes of any
one-node marginals computed by BP converge to the average $\overline{y}$ as $\beta$ tends to 
infinity (as follows from~\cite{MJW2006}).
In this way estimates of $\overline{y}$ can be obtained, where a trade-off must be made between 
convergence time and accuracy.

\paragraph*{The algorithm --}
The following message update rules define Consensus Propagation:
\begin{align}
K_{ij}^{(t+1)} &=\frac{1+\sum_{k \in N(i)\backslash j}K_{ki}^{(t)}}{1+\frac{1}{\beta Q_{ij}}(1+\sum_{k \in N(i)\backslash j}K_{ki}^{(t)})} \label{eqn:K-message} \\
\mu_{ij}^{(t+1)} &=\frac{y_i+\sum_{k \in N(i)\backslash j}K_{ki}^{(t)}\mu_{ki}^{(t)}}{1+\sum_{k \in N(i)\backslash j}K_{ki}^{(t)}} \label{eqn:mu-message}
\end{align}
This parametrization of BP yields two-dimensional real-valued messages consisting of a \textit{topology message} $K$ and a \textit{local state update} $\mu$. The notation $X_{ij}^{(t)}$ means that message $X$ is sent from originating node $i$ to target node $j$ at iteration step $t$: $N(i)$ is the set of all neighbors of node $i$, and  $N(i)\backslash j$ is the set of all neighbours of $i$ except $j$. The algorithm is said to have attained \textit{consensus}, if the messages are fixed points of \eqref{eqn:K-message} and \eqref{eqn:mu-message}. A belief for the average $\overline{y}$ at time $t$ and node $i$ is obtained via the CP output rule:
\begin{equation}
\label{eqn:y_average}
\overline y_i = \frac{y_i+\sum_{k \in N(i)}K_{ki}^{(t)}\mu_{ki}^{(t)}}{1+\sum_{k \in N(i)}K_{ki}^{(t)}}
\end{equation}
The consensus beliefs \eqref{eqn:y_average}, with $K$ and $\mu$ at a fixed point
of \eqref{eqn:K-message} and \eqref{eqn:mu-message}, are the Belief Propagation predictions of
the modes of the one-node marginals of the probability distribution \eqref{eqn:CP}.

\begin{figure}
\includegraphics[width=0.30\textwidth]{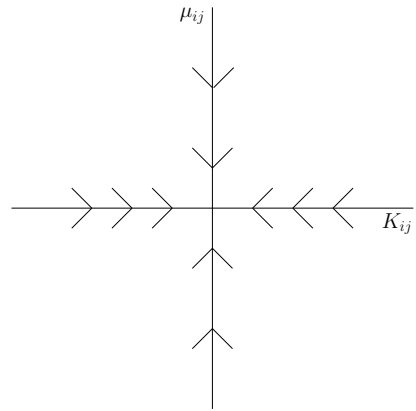}
\caption{Convergence scheme for Consensus Propagation. The $K$-subspace is a fast, the $\mu$-subspace a slow stable manifold.}
\label{fig:convergence_scheme}
\end{figure}

\paragraph*{Convergence for different initializations --}
\begin{figure}
\centering
\includegraphics[width=0.50\textwidth]{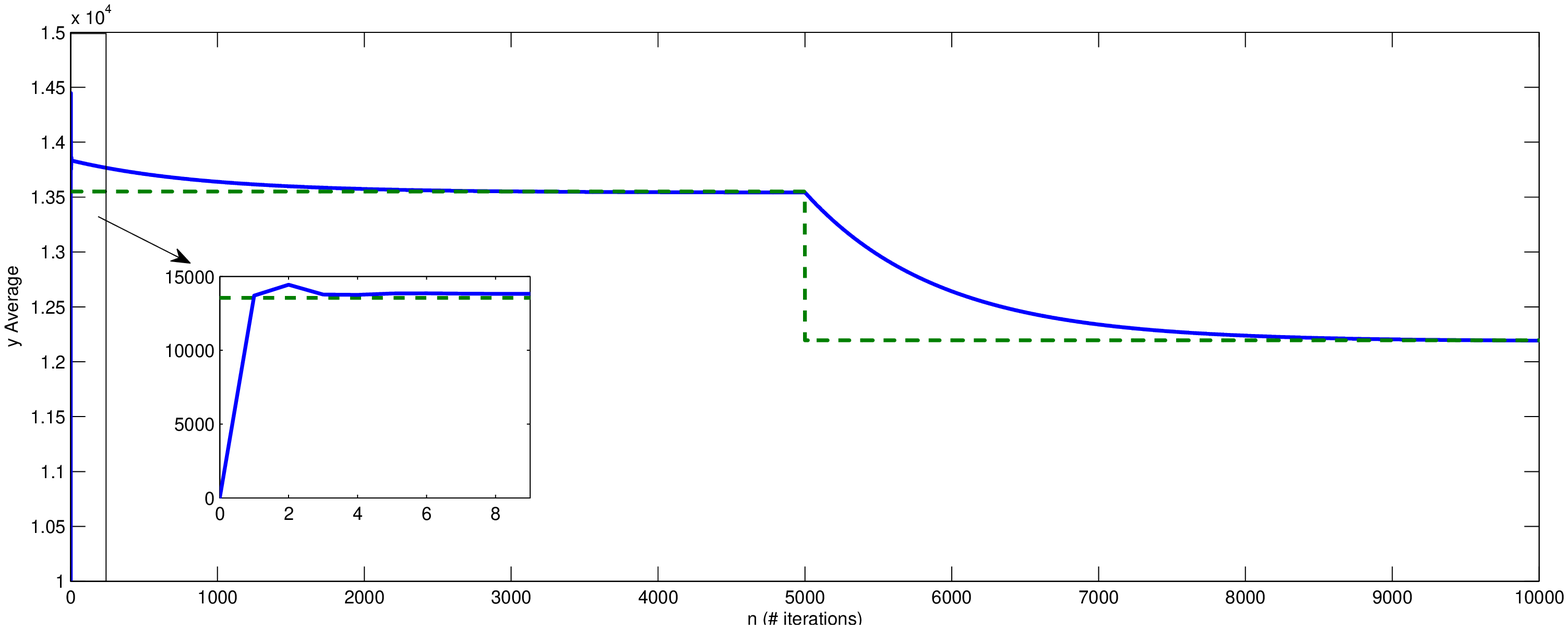}
\caption{Convergence of the $\overline{y}$ belief at one node in a random network with 500 nodes. The solid line is the CP performance, the dashed line indicates the correct y-average. Insert shows behaviour in 1-10 rounds; main figure shows behaviour in $10^3-10^4$ rounds. Node values were generated randomly, and then 
scaled by $90\%$ in round $5*10^3$. At iteration step $n=0$ all CP-messages were 
initialized uniformly to zero. Fast convergence followed by an overshoot (damped oscillation)
is observed. After the reset at $n=5*10^3$ the CP-messages were left unchanged, and a markedly slower
convergence, but without an overshoot, is observed.}
\label{fig:CP_500_1}
\end{figure}
Figure \ref{fig:CP_500_1} shows performance of Consensus Propagation after initializing all messages to zero: the algorithm shows first an oscillating behaviour with \textit{fast} convergence to a good approximation of the correct mean $\overline{y}$. After changing every node value and NOT re-initializing the messages, the algorithm exhibits a steady, yet much \textit{slower}, convergence. This second behaviour corresponds to the case of \textit{dynamic data}, where the topology message ($K$-message) and local state update ($\mu$-message) start at their converged values before the perturbation. Once a fixed point $K^*$ is reached, the topology messages will not change if only the local measurement values $y_i$ change, since \eqref{eqn:K-message} is an equation only involving topology messages. Except for an initial transient, the dynamic data case can hence be completely understood by the linear operator expressed by the right-hand side of \eqref{eqn:mu-message} (see below). Generally, it seems that the topology messages converge much faster than the local state messages, and that therefore the linear theory (explained below) also bounds the behaviour of dynamic data, where both local values $y_i$ and local couplings $Q_{ij}$ change. Before we turn to the linear analysis, let us however point out the observation that different initializations of the messages yield different performance, and, perhaps surprisingly, that initializing with $K=\mu=0$ seems to be the superior choice.
\begin{figure}
\centering
\includegraphics[width=0.50\textwidth]{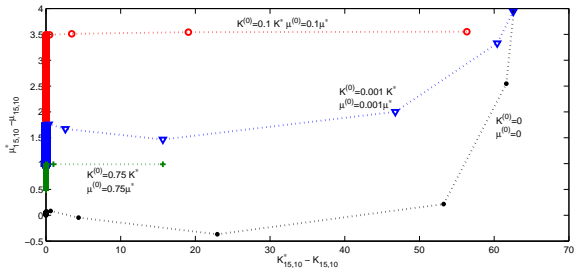}
\caption{Convergence behaviour of Consensus Propagation on a random graph for different initial messages. As model we used a random graph with 80 nodes, all edges present with  probability 0.1, $\beta=100$ and $Q_{ij}$ chosen i.i.d. random variables uniform between $0.5$ and $2$. The plot shows the time evolution of the deviation of two messages from their converged values ($K^\ast-K$ and $\mu^\ast -\mu$), sent from node 15 to node 10 during 500 iterations. The messages were initialized proportional to their fixed point values, and start at the top right corner of each trajectories in the figure. The trajectories exhibit an initial fast decay of the error in topology messages (abscissa) followed by a slower decay of the local state message (ordinate). The only exception is when the messages are initialized as  $K=\mu=0$ in which case the trajectory seems to fall into the (more) stable manifold of the fixed point (a ``direct hit''), with the second slow process along the ordinate absent. For a graphical illustration of the conjectured behaviour, see \protect\ref{fig:convergence_scheme}. The fixed point in this example have $K_{15,10}^\ast=62.61$ and $\mu_{15,10}^\ast=3.95$.}
\label{fig:initialization}
\end{figure}
The observations of Fig. \ref{fig:initialization} contradict a conjecture put forward in~\cite{MoaVan2006} that convergence times for $K^{(0)}=0$ and $K^{(0)}=K^\ast$ are equivalent. In fact, initializing with $K=0$ improves convergence dramatically. Let us note that if $K$ be re-initialized to zero, then the re-initialization of $\mu$ is arbitrary, since by (\ref{eqn:mu-message}), $\mu^{(1)}$ will then be equal to $y_i$, \textit{i.e} independent of $\mu^{(0)}$. In a scenario where many measurement values (and/or also the underlying network) change simultaneously, re-starting Consensus Propagation using $K=0$ may therefore by a valid option. We stress that this is not obvious, but follows if the dynamic behaviour is as illustrated in Fig. \ref{fig:convergence_scheme}. This may not be true in all underlying topologies. However, in the case that the underlying topology is locally tree-like, as is the case for the random graphs in Fig. \ref{fig:initialization}, a heuristic explanation for the faster convergence of Consensus Propagation, when initializing with $K=0$, is the following: as was shown by \cite{MoaVan2006} Consensus Propagation yields the exact node average on tree-like graphs with the global coupling constant $\beta=\infty$ and $K^{(0)}=0$. Initializing CP on a random graph with $K=0$ and a large value of $\beta$ will yield nearly the same messages, after a finite number of iterations, as initializing CP with $K=0$ on a computational tree associated with the graph (using the construction of~\cite{MJW2006}) at $\beta=\infty$. This explains the improved behaviour of starting with $K=0$ \textit{qualitatively}, but does not explain it \textit{quantitatively}, \textit{i.e.} the apparent complete absence of the slow process in Fig. \ref{fig:initialization}.

\paragraph*{Theory of Consensus Propagation --}
Consensus Propagation can be considered as non-linear dynamical system in a multidimensional space spanned by all $K$- and $\mu$- messages:
\begin{equation}
\begin{pmatrix}
\mu_{ij}^{(n+1)}\\
...\\
K_{ij}^{(n+1)}\\
...
\end{pmatrix}
=\mathbb{R}
\begin{pmatrix}
\mu_{ij}^{(n)}\\
...\\
K_{ij}^{(n)}\\
...
\end{pmatrix}
\end{equation}

The numerical experiments above indicate that the $\mu$-message subspace spans a slow stable and the $K$-message subspace a fast stable manifold (see Fig. \ref{fig:convergence_scheme}). We will use the eigenvalues of a linearized version of $\mathbb{R}$ to verify this. Following \cite{MJW2006}, we refer to the non-linear iterated map transfer operator $\mathbb{R}$ as a \textit{Ruelle-Peron-Frobenius Operator}. The linear part of $\mathbb{R}$ has the matrix representation:
\begin{align}
\begin{pmatrix}
\mu_{ij}^{(n+1)}\\
...\\
K_{ij}^{(n+1)}\\
...
\end{pmatrix}&=
\begin{pmatrix}
\mu_{ij}^\ast + \Delta \mu_{ij}^{(n+1)}\\
...\\
K_{ij}^\ast + \Delta K_{ij}^{(n+1)}\\
...
\end{pmatrix}\\
&=
\begin{pmatrix}
\mu_{ij}^\ast\\
...\\
K_{ij}^\ast\\
...
\end{pmatrix}+
\mathbb{R}'
\begin{pmatrix}
\Delta \mu_{ij}^{(n)}\\
...\\
\Delta K_{ij}^{(n)}\\
...
\end{pmatrix}
\end{align}
$\mathbb{R}'$ is the linearized transfer operator. The matrix representation of this operator can be decomposed into four quadratic submatrices:
\begin{equation}
R'=
\begin{pmatrix}
A&C\\
0&B
\end{pmatrix}
\label{eqn:Rprime-def}
\end{equation}
\begin{figure}
\includegraphics[width=0.50\textwidth]{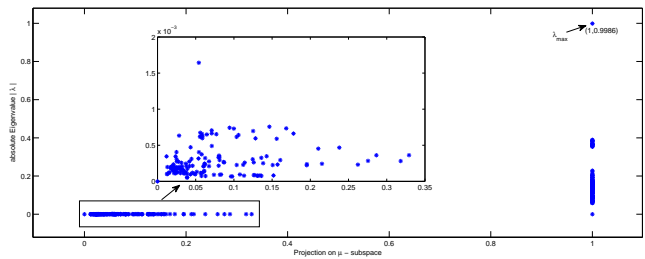}
\caption{Length of projections of (normlength) eigenvectors on the $\mu$-subspace for each eigenvalue $\lambda$ of the linearized Ruelle-Peron-Frobenius-Operator $\mathbb{R}'$ for a $G(N=20,c=8)$ Erd\H{o}s-R\'{e}nyi model.
$\beta=100$, and $Q_{ij}$ randomly generated as in Fig~\protect\ref{fig:initialization}.}
\label{fig:proj2}
\end{figure}
Submatrix $A$ is the transfer matrix in the \textit{dynamic data} case, when the topology messages have converged, submatrix $B$ is the linear part of the transfer matrix acting in the \textit{dynamic network} on the topology messages alone, and submatrix $C$ is the linear action of the topology messages on the local state variables. Around the fixed point, we can verify that topology messages converge faster than local state updates, by comparing the size of the eigenvalues of $\mathbb{R}'$ to the projection of the corresponding eigenvectors on the subspace spanned by the $\mu$-messages. As shown in Fig. \ref{fig:proj2}, the (isolated) largest eigenvalue lies in the subspace of local state updates. In addition, most of the other eigenvectors in the subspace of local updates also have eigenvalues larger than all the eigenvalues projecting on the topology messages. 
Table~\ref{tab:lambda_A_B} compares the leading eigenvalues of submatrices $A$ and $B$
for four different Erd\H{o}s-R\'{e}nyi graphs, reinforcing the observation from
Fig. \ref{fig:proj2}.
\begin{table}
\begin{tabular}{|c|c|c|c|}
\hline
N&c&$\lambda_{max}(A)$&$\lambda_{max}(B)$\\ \hline
20&18&0.99949152&0.00054415\\ \hline
30&14&0.99924356&0.00083034\\ \hline
40&10&0.99895415&0.00119833\\ \hline
50&8&0.99851962&0.00186674\\ \hline
\end{tabular}
\caption{Comparison of leading eigenvalues of linearized matrices $A$ and $B$ in four Erd\H{o}s-R\'{e}nyi graphs 
$G(N,p=c/N)$. The much smaller eigenvalues of matrix $B$ imply much faster convergence of the topology message
s.}
\label{tab:lambda_A_B}
\end{table}
In linear theory, the limiting factor on convergence is therefore the dynamics of the local state updates.

\begin{figure}[h]
\centering
\includegraphics[width=0.50\textwidth]{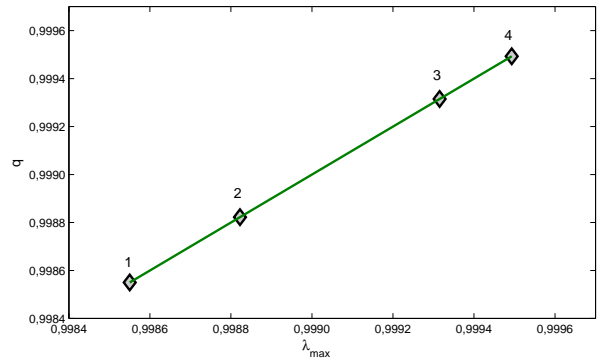}
\caption{Convergence ratios of the linear averaging process \protect\eqref{eqn:linearized-averaging} compared to leading eigenvalue of operator $A$. and numerical calculated leading eigenvalues $\lambda_{max}$ in four examples of  Erd\H{o}s-R\'{e}nyi models $G(N,p=c/N)$. $q$: convergence ration, $\lambda_{max}$: largest eigenvalue. The solid line represents $q=\lambda_{max}$. Cases: 1: $c=8$, $N=50$, 2: $c=10$, $N=40$, 3: $c=14$, $N=30$, 4: $c=18$, $N=20$.
$\beta=100$, and $Q_{ij}$ randomly generated as in Fig~\protect\ref{fig:initialization} for all cases.}
\label{fig:q_lambda_plot}
\end{figure}

\paragraph*{Dynamic data case --}The case when topology messages have converged is also of interest when data to be measured keep on changing: in this scenario CP is a \textit{linear averaging process}. Indeed, the local state update equation (\ref{eqn:mu-message}) is then a linear equation of one free vector $\mathbb{\mu}=(...\mu_{ij}..)$ and can be expressed in linear operator form:
\begin{equation}
\boldsymbol{\mu}^{(n+1)}=\boldsymbol{b}+A\boldsymbol{\mu}^{(n)}
\label{eqn:linearized-averaging}
\end{equation}
The operator $A$ in \eqref{eqn:linearized-averaging} (acting on $\mu$-messages)
is the same as the submatrix $A$ of the operator $\mathbb{R}'$ in \eqref{eqn:Rprime-def},
and its spectral properties are as described in Fig.~\ref{fig:proj2} (rightmost set
of eigenvalues, all completely in the subspace of local state updates).
The vector $\boldsymbol{b}$ is the $\mu$-independent part of \eqref{eqn:mu-message},
and in particular depends on the set of local measurement values $\boldsymbol{y}$.
If these change in time, \eqref{eqn:linearized-averaging} is obviously a linear 
averaging process with kernel $A$. If the  $\boldsymbol{y}$ do not change, and the
$\mu$-messages are initialized in some manner, we expect from Fig.~\ref{fig:proj2} 
that convergence will eventually be dominated by the largest (isolated) eigenvalue of $A$.
Fig.~\ref{fig:q_lambda_plot} shows that this is indeed the case, for several different 
Erd\H{o}s-R\'{e}nyi graphs. In these models, we always find an isolated largest eigenvalue
(data not shown).

\paragraph*{Scalability of CP in Erd\H{o}s-R\'{e}nyi graphs --}
The above discussion leads up to the conclusion that the largest eigenvalue of operator $A$ 
of \eqref{eqn:linearized-averaging} is a quantity of major importance to understand
the performance of CP in dynamic environments -- both dynamic data only, and also dynamic
network. The scaling properties of this largest eigenvalue therefore determines
how effective the CP averaging procedure can be in a large network.
Following the general principles of random graph theory, we should
compare random graphs of increasing size $N$, but with the same average node degree $c$.
This means that every link is present in the graph with probability
$p=\frac{c}{N}$ (up to corrections decaying with $N$).
Table~\ref{tab:E-R-test1} shows that in  
a family of Erd\H{o}s-R\'{e}nyi graphs
with asymptotic average node degree $c=8$ the largest of eigenvalue $A$
seems to converge to a finite value less than one.
In the experiments, the local couplings $Q_{ij}$ are generated randomly
between $0.5$ and $2$. The fifth column gives (for the smaller instances)
the standard deviation of the largest eigenvalue computed from $100$
experiments (independent realizations of the random graphs, and independent
realizations of the local coupling constants $Q_{ij}$). The decay of
the standard deviation with $N$ indicates that the leading eigenvalue is
a self-averaging quantity in this ensemble.
\begin{table}[h]
\begin{tabular}{|c|c|c|c|c|c|c|}
\hline
N&p&$c_{exp}$&$[c_{exp}]_{100}$&$\sigma [c_{exp}]_{100}$&$\lambda_{max}$&$[\lambda_{max}]_{100}$\\ \hline
20&0.4&7.3&7.6&0.6&0.9984&0.9985\\ \hline
40&0.2&7.7&7.8&0.5&0.9985&0.9985\\ \hline
80&0.1&7.8&7.9&0.4&0.9986&0.9985\\ \hline
160&0.05&8.2&8.0&0.3&0.9987&0.9985\\ \hline
5000&0.0016& & & &0.99850&\\ \hline
10000&0.0008& & & &0.99851&\\ \hline
20000&0.0004& & & &0.99850&\\ \hline
\end{tabular}
\caption{Convergence ratios for Erd\H{o}s-R\'{e}nyi graphs. Coupling constant $\beta=100$ and $Q_{ij}$ were chosen randomly uniform between $0.5$ and $2$. All instances have a theoretical average node degree $c=8$. The table shows the outcome of a single experiment ($c_{exp}$, $\lambda_{max}$) and for small graphs of 100 experiments ($[c_{exp}]_{100}$, $\sigma [c_{exp}]_{100}$, $[\lambda_{max}]_{100}$).}
\label{tab:E-R-test1}
\end{table}
Let us note that our results concur with (and extend) a result
of \cite{MoaVan2006} for regular graphs, where the authors showed that convergence time 
is not dependent on the graph size. If this be true, the leading eigenvalue in that
ensemble must also be a self-averaging quantity, independent of graph size.\\

\begin{figure}[h]
\centering
\includegraphics[width=0.50\textwidth]{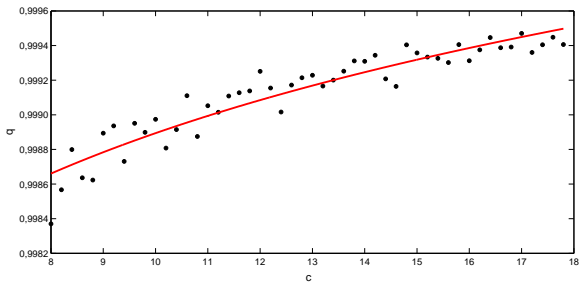}
\caption{Dependence of convergence ratio $q$ on average node degree $c$ in an Erd\H{o}s-R\'{e}nyi graph with 20 nodes. The solid line is a data fit: q= $0.001046 \cdot \log (c)+0.9965$}
\label{fig:E-R-test2}
\end{figure}

Fig.~\ref{fig:E-R-test2} shows the dependence of the leading eigenvalue on the node degree $c$, for a number of graphs with $20$ nodes. The eigenvalue shows an increasing trend, in this range fairly well approximated by a logarithmic behaviour. 
 
\paragraph*{Summary --}
Statistical physics has contributed very significantly in recent years to the understanding of Belief Propagation approaches to \textit{inference}, which have very important applications to \textit{e.g.} iterative decoding~\cite{MezardMontanari2009}. In this contribution, we have looked at a Belief Propagation-based algorithm for \textit{averaging}, with potentially numerous applications to network management. We showed that this Consensus Propagation algorithm, in a dynamic environment, is a dynamical system which can be fruitfully analysed by the tools of statistical physics and nonlinear dynamics. We showed that CP responds quickly to changes in the network topology, and more slowly to changing data. This can be understood intuitively as a dynamic network improves mixability, which should not be a disadvantage when computing an average (or an estimate of an average). In a real world application, CP is therefore not limited by a changing network structure but by dynamic data. 
Secondly, and of interest to statistical physicists, we exhibited an interesting self-averaging property of the leading eigenvalue of the transfer matrix describing the the dynamic data case. Perhaps surprisingly, this leading eigenvalue seems to be asymptotically independent of network size.

\paragraph*{Acknowledgement --}
R.P. was partially funded by the European "Life Long Learning Program" under project number DE-2008-ERA/MOB-KonsZuV01-CP07. E.A. acknowledges support from the Swedish Science Council through the KTH Linnaeus Centre ACCESS, and from the Academy of Finland.


\begin{thebibliography}{so}

\bibitem{Pearl1988} Pearl, "Probabilistic Reasoning In Intelligent Systems", Morgan Kaufmann, 1988 

\bibitem{Cow1998} Cowell, "Advanced Inference in Bayesian Networks", in "Learning in Graphical Models", edited by Michael Jordan, 1998 Kluwer Academic Publishers

\bibitem{Kschischang2001} Kschischang, Frey and Loeliger, ``Factor Graphs and the Sum-Product Algorithm'' IEEE Transactions on Information Theory, vol. 47, pp. 498-519, February 2001

\bibitem{YedFreWei2003} Yedidia, Freeman and Weiss, "Understanding Belief Propagation And Its Generalization" in "Exploring Artificial Intelligence in the New Millennium", (Science \& Technology Books, TR2001-022), January 2003

\bibitem{MezardMontanari2009} Mezard and Montanari, "Information, Physics, and Computation", Oxford University Press, 2009 

\bibitem{MoaVan2006} Moallemi and Van Roy, "Consensus Propagation", IEEE Transactions on Information Theory, Vol. 52, No. 11, November 2006 

\bibitem{JeMoBa2005} Jelasity, Montresor, and Babaoglu
``Gossip-based aggregation in large dynamic networks'',
 ACM Transactions on Computer Systems, 23(3):219-252, August 2005.

\bibitem{WeFree2001} Weiss and Freeman
``Correctness of Belief Propagation in Gaussian graphical models of arbitrary topology'', 
Neural Computation 13:2173-2200 (2001)

\bibitem{MJW2006} Malioutov, Johnson and Willsky, "Walk-Sums And Belief Propagation In Gaussian Graphical Models", Journal of Machine Learning Research 7 (2006), pp. 2031-2064

\bibitem{WeiFre2001} Weiss and Freeman, "Correctness Of Belief Propagation In Gaussian Graphical Models Of Arbitrary Topology" Neural Computation, Vol. 13, No. 10, pp. 2173-2200, October 2001

\bibitem{MurWeiJo} Murphy, Weiss and Jordan, "Loopy Belief Propagation For Approximate Inference: An Empirical Study", Proceedings of the 15th Annual Conference on Uncertainty in Artificial Intelligence (UAI-99), Morgan Freeman, pp. 467-47, 1999


\end{thebibliography}
\end{document}